%% file: main.tex
\documentclass[11pt]{article}
\usepackage{amsfonts}
\usepackage{amsmath}
\usepackage{amssymb}
\usepackage{array}
\usepackage{authblk}
\usepackage{booktabs}
\usepackage{caption}
\usepackage{gensymb}
\usepackage[margin=1in]{geometry}
\usepackage{graphicx}
\usepackage{hhline}
\usepackage{makecell}
\usepackage{multirow}
\usepackage{scalerel}
\usepackage{subcaption}
\usepackage{xcolor}
\usepackage[colorlinks=true,citecolor=blue]{hyperref}
\usepackage[numbers,sort&compress]{natbib}

\newcolumntype{P}[1]{>{\centering\arraybackslash}p{#1}}
\newcolumntype{M}[1]{>{\centering\arraybackslash}m{#1}}

\title{Efficient and generalizable nested Fourier-DeepONet for three-dimensional geological carbon sequestration}

\author[1,\dag]{Jonathan E. Lee}
\author[2,\dag]{Min Zhu}
\author[3]{Ziqiao Xi}
\author[4]{Kun Wang}
\author[4,*]{Yanhua O. Yuan}
\author[2,*]{Lu Lu}

\affil[1]{Department of Chemical and Environmental Engineering, Yale University, New Haven, CT 06511, USA}
\affil[2]{Department of Statistics and Data Science, Yale University, New Haven, CT 06511, USA}
\affil[3]{Department of Computer Science and Engineering, University of California, San Diego, CA 92093, USA}
\affil[4]{ExxonMobil Technology and Engineering Company, Annandale, NJ 08801, USA}
\affil[$\dag$]{These authors contributed equally to this work.}
\affil[*]{Corresponding author. Email: yanhua.yuan@exxonmobil.com, lu.lu@yale.edu}

\date{}

\begin{document}
\maketitle

\begin{abstract}
Geological carbon sequestration (GCS) involves injecting CO\textsubscript{2} into subsurface geological formations for permanent storage. Numerical simulations could guide decisions in GCS projects by predicting CO\textsubscript{2} migration pathways and the pressure distribution in storage formation. However, these simulations are often computationally expensive due to highly coupled physics and large spatial-temporal simulation domains. Surrogate modeling with data-driven machine learning has become a promising alternative to accelerate physics-based simulations. Among these, the Fourier neural operator (FNO) has been applied to three-dimensional synthetic subsurface models. Here, to further improve performance, we have developed a nested Fourier-DeepONet by combining the expressiveness of the FNO with the modularity of a deep operator network (DeepONet). This new framework is twice as efficient as a nested FNO for training and has at least 80\% lower GPU memory requirement due to its flexibility to treat temporal coordinates separately. These performance improvements are achieved without compromising prediction accuracy. In addition, the generalization and extrapolation ability of nested Fourier-DeepONet beyond the training range has been thoroughly evaluated. Nested Fourier-DeepONet outperformed the nested FNO for extrapolation in time with more than 50\% reduced error. It also exhibited good extrapolation accuracy beyond the training range in terms of reservoir properties, number of wells, and injection rate.
\end{abstract}

\paragraph{Keywords:} geological carbon sequestration; 3D multiphase flow in porous media; deep neural operator; Fourier-DeepONet; computational cost; extrapolation

\input{content/introduction}
\input{content/problemsetup}

\input{content/methods}
\input{content/results}
\input{content/conclusions}

\section*{Acknowledgments}

This work was supported by ExxonMobil Technology and Engineering Company and U.S. Department of Energy Office of Advanced Scientific Computing Research under Grants No.~DE-SC0025592 and No.~DE-SC0025593. We thank Gege Wen et al. for making the dataset public and Alex GK Lee for providing valuable feedback.

\section*{Disclosure statement}

No potential conflict of interest was reported by the authors.

\section*{Author contributions statement}

\textbf{Jonathan E. Lee:} Writing – review \& editing, Writing – original draft, Visualization, Validation, Software, Methodology, Investigation, Formal analysis, Data curation. \textbf{Min Zhu:} Writing – review \& editing, Writing – original draft, Visualization, Validation, Software, Methodology, Investigation, Formal analysis, Data curation. \textbf{Ziqiao Xi:} Visualization, Investigation. \textbf{Kun Wang:} Writing – review \& editing, Supervision, Funding acquisition. \textbf{Yanhua O. Yuan:} Writing – review \& editing, Supervision, Funding acquisition. \textbf{Lu Lu:} Writing – review \& editing, Writing – original draft, Supervision, Resources, Project administration, Methodology, Funding acquisition, Formal analysis, Conceptualization.

\section*{Data availability statement}

Training data and the code supporting the findings of this study will be available in GitHub at \url{https://github.com/lu-group/nested-fourier-deeponet-gcs-3d}. These data were derived from the following resources available in the public domain: Ref.~\cite{wen2023real}.

\appendix
\input{content/appendix}
\bibliographystyle{acm}
\bibliography{main}

\end{document}

%% file: content/introduction.tex
\section{Introduction}
\label{sec1} 


Geological carbon sequestration (GCS) is an essential technology to reduce CO\textsubscript{2} emissions at scale. High-fidelity simulations of plume migration pathways and pressure distribution during and after CO\textsubscript{2} injection play an important role in decision making in carbon storage projects. The problem can be modeled by solving a multiphase flow of CO\textsubscript{2} and water, governed by Darcy's law, which is expressed by nonlinear partial differential equations (PDEs)~\cite{Pruess2002,Blunt2016}. Numerical simulations, such as finite difference, finite element, or finite volume methods, provide insights to guide informed decisions about subsurface scenarios. These simulations allow the prediction of how much CO\textsubscript{2} will migrate through sedimentary rocks for effective site selection and injection designs. However, this approach is computationally demanding for performing optimization tasks or decision-making processes, due to not only the multiphysics, multiscale, and multiphase nature of this problem but also the need for high grid resolution~\cite{Pruess2002, Blunt2016, Orr2007, Doughty2009, Wen2019, Khebzegga2020}. As a result, the trade-off between accuracy and computational cost is inherent in such simulation methods. One technique to address this issue involves subdividing the grid into multiple resolutions, thereby reducing overall computational costs. Local grid refinement (LGR) technique is typically employed to maintain solver accuracy near the CO\textsubscript{2} injection location~\cite{wen2023real}. This method discretizes the region around injection wells with finer grids and gradually coarsens regions further away from them~\cite{Bramble1988, Eigestad2009, Faigle2014, Kamashev2021}. This approach enables the simulation model to capture the complex physics of CO\textsubscript{2} plume migration around the wells with minimal loss of accuracy. However, it is still computationally expensive to use such techniques to model real-world scenarios.

Machine learning (ML) methods have emerged as a potential alternative to traditional numerical approaches due to their capability to learn PDEs and make real-time predictions~\cite{Chen2020, Yazdani2020, Daneker, Zhu2018, Mo2019, Tang2020, Wen2021ccsnet, jiao2021one}. Recent advances in ML have shown significant progress in solving subsurface flow problems~\cite{Wen2019, wen2023real, Zhu2018, Mo2019, Tang2020, Wen2021ccsnet, Wen2021plume}. ML techniques offer rapid inference, making them particularly useful for tasks such as probabilistic assessment, which require multiple forward simulations. Notably, physics-informed neural networks (PINNs) and their extensions~\cite{Karniadakis2021, jagtap2020extended, Yu2022, Lu2021_PINN} have gained popularity from their ability to make informed predictions in various scientific and engineering applications and already have shown success in many applications~\cite{Chen2020, Yazdani2020, Daneker, fan2023deep, Daneker2024, Wu2023, Hayford2024, Wu2023residual}. Despite their successes, PINNs struggle with computational cost in real-world scenarios due to the necessity for separate training if DE conditions vary. For example, in GCS, various reservoir conditions and varying injection schemes change the CO\textsubscript{2} migration behavior, which makes PINNs inefficient for this application. 

Recently, deep neural operators have gained significant attention because of their generalizability in learning the PDE family rather than a single PDE with fixed parameters. They do not require separate training for new conditions~\cite{Lu2021_DEEPONET, Li2020, Lu2022, zhu2023reliable, zhu2023fourier, jiang2024fourier, Mao2023, Deng2022, Lu2022multifidel, Zhang2024, ClarkDiLeoni2023, Jin2022}. Furthermore, in the field of GCS, a nested Fourier neural operator (FNO) framework has been used to simulate pressure buildup and carbon saturation in three-dimensional (3D) spatial and temporal domains, mirroring real-world scenarios~\cite{wen2023real}. Nested FNO adopts a similar idea as LGR technique using a separate model per each level of resolution. Despite using this technique, due to the large temporal and 3D spatial domain it still requires a couple of days of training for each level of resolution, and a GPU with large memory ($>$30 GiB). Furthermore, because FNO uses different channels to predict solutions at different times, it performs poorly in predicting unseen times~\cite{jiang2024fourier}.

In our study, we aim to address the challenges arising from computational cost and poor generalizability. We develop a nested Fourier-DeepONet by employing a deep neural operator called Fourier-DeepONet~\cite{zhu2023fourier,jiang2024fourier}, which integrates the cutting-edge FNO~\cite{Li2020} with the deep operator network (DeepONet)~\cite{Lu2021_DEEPONET}. Nested Fourier-DeepONet offers significant improvements in GPU memory usage and training time compared to the state-of-the-art nested FNO~\cite{wen2023real} owing to its design of incorporating separate networks for handling temporal coordinates. This nificantly increases computational efficiency with at least 50\% reduction in GPU memory consumption and 50\% decrease in training time. Additionally, a nested Fourier-DeepONet demonstrates promising results in extrapolation in terms of sampling parameters that include various injection schemes, subsurface conditions, and temporal coordinates. 

The paper is organized as follows. In Section~\ref{sec:problemsetup}, we introduce the problem setup, including the governing equations of multiphase flow and the dataset generated by numerical simulation. In Section~\ref{sec:methods}, after providing a brief overview of DeepONet and FNO, we propose a nested Fourier-DeepONet. In Section~\ref{sec:results}, we highlight the superior computational efficiency of the nested Fourier-DeepONet compared to the nested FNO and demonstrate its generalization capabilities. We summarize the significance of this work in Section~\ref{sec:conclusion}. 

%% file: content/problemsetup.tex
\section{Problem setup}\label{sec:problemsetup}

In this work, we investigate carbon migration in varying scenarios of injection schemes and subsurface conditions for GCS. We model 3D GCS as a multiphase flow of CO\textsubscript{2} and water in a porous media to predict pressure buildup and gas saturation when injecting CO\textsubscript{2} underground over 30 years. To alleviate the computational cost arising from the large spatial and temporal domain, the local refinement technique is employed for data generation, and a nested ML framework is constructed accordingly to make predictions at varying discretizations. 

\subsection{Governing equations}\label{sec2.1}
The problem of GCS is governed by mass balance equations for CO\textsubscript{2} and water, where the flux of each phase is determined by an extension of Darcy’s law, which describes fluid flow through a porous medium~\cite{Pruess2002} as follows:
\begin{equation}
    \frac{\partial\left(\phi\sum_{p}S_{p}\rho_{p}X^{CO_{2}}_{p}\right)}{\partial t} = -\nabla \cdot \left[\mathbf{F}^{CO_2}|_{adv} + \mathbf{F}^{CO_2}|_{dif}\right] + q^{CO_2},
\label{eq1}
\end{equation}
\begin{equation}
    \frac{\partial\left(\phi\sum_{p}S_{p}\rho_{p}X^{water}_{p}\right)}{\partial t} = -\nabla \cdot \left[\mathbf{F}^{water}|_{adv} + \mathbf{F}^{water}|_{dif}\right],
\label{eq2}
\end{equation}
\begin{equation*}
    \mathbf{F}^{\eta}|_{adv} = \sum_{p} X^{\eta} F_p = \sum_{p} X^{\eta} \left(-k \frac{k_{r,p} \rho_{p}}{\mu_{p}}(\nabla P_{p} - \rho_{p} \mathbf{g})\right),
\end{equation*}
\begin{equation*}
    P_{n} = P_{w} + P_{c}.
\end{equation*}
Eqs.~\eqref{eq1} and~\eqref{eq2} represent the law of conservation of mass for the two species within the system: CO\textsubscript{2} and water. On the left side of the equations, $\phi$ denotes the porosity of the media, $S$ represents the saturation, $\rho$ stands for density, and $X$ signifies the mass fraction of each species, with the subscript $p$ indicating the phase. Lastly, $q^{CO_2}$ denotes the injection rate of CO\textsubscript{2}.

Although molecular diffusion and hydrodynamic dispersion, $\textbf{F}_{dif}$, are not explicitly simulated, they are implicitly considered as part of numerical diffusion and dispersion resulting from finite difference gradient approximation~\cite{wen2023real}. The advective mass flux, denoted as $\textbf{F}_{adv}$, for component $\eta$, is governed by Darcy’s law for each phase. Here, $k$ denotes absolute permeability, $P$ stands for fluid pressure, $\textbf{g}$ represents gravitational acceleration, $k_{r}$ signifies relative permeability, and $\mu$ indicates viscosity. The fluid pressure for the non-wetting phase, $P_{n}$, is the sum of the wetting phase pressure, $P_{w}$, and the capillary pressure, $P_{c}$.

\subsection{Nested grid refinement framework}\label{sec2.2}

\begin{figure}[htbp]
    \centering
        \includegraphics[width=\textwidth]{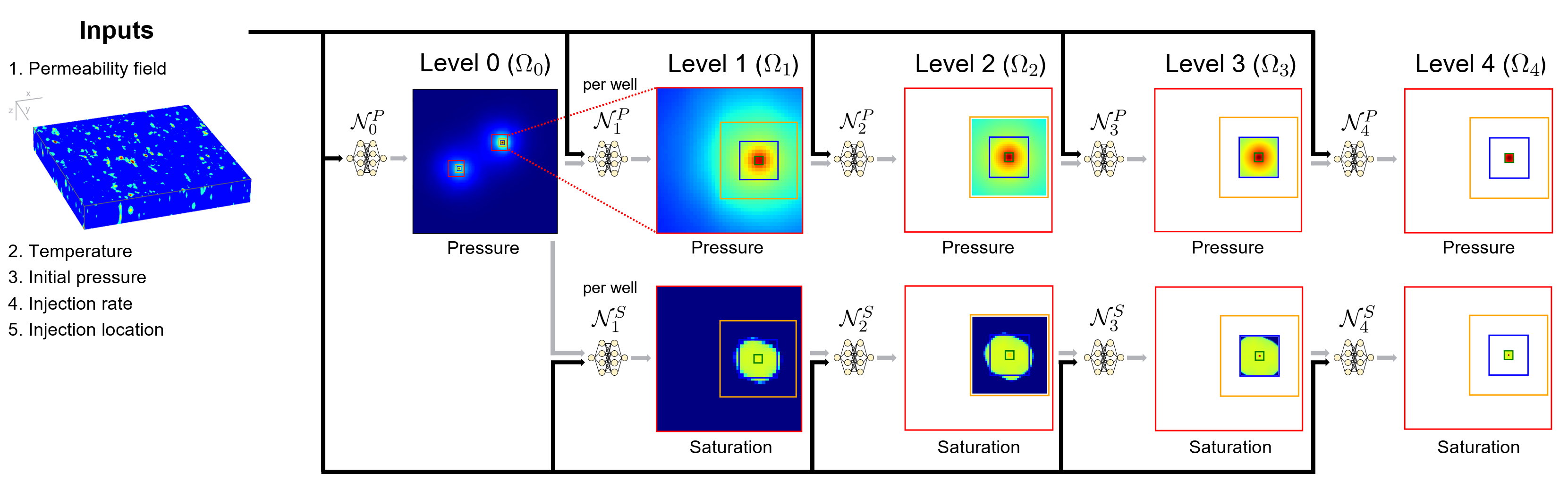}
    \caption{\textbf{Nested machine learning pipeline for predicting the pressure buildup and gas saturation for five levels with gradually increasing resolution.} There are two groups of inputs which include reservoir condition (3D permeability field, temperature, 3D initial pressure) and injection scheme (injection rate, and injection location). Each level starting from level 1 uses outputs from the previous level (grey arrows) on top of the reservoir condition and injection scheme (black arrows).}
    \label{fig:nested}
\end{figure}

In this work, we aim to predict pressure buildup and gas saturation over a 30-year period based on reservoir conditions (i.e., permeability field, temperature, and initial pressure) and the injection scheme (i.e., injection rate and location) (Fig.~\ref{fig:nested}). 

\begin{figure}[htbp]
    \centering
        \includegraphics[width=\textwidth]{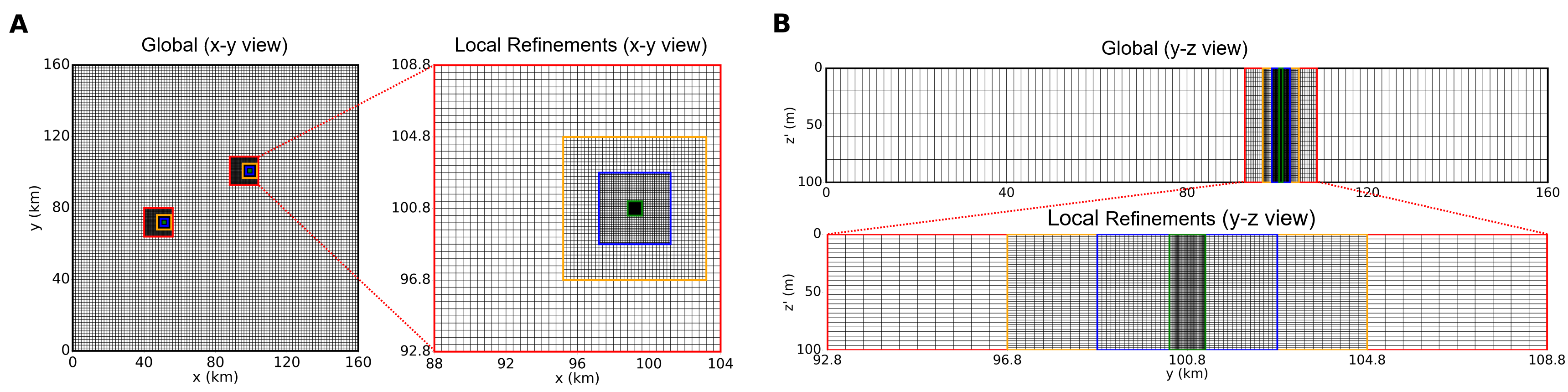}%
    \caption{\textbf{2-Dimensional sample visualizations of a global grid and local refinements in the dataset.} (\textbf{A}) $x$-$y$ view. (\textbf{B}) $y$-$z$ view. The boundaries of different computational domains are drawn in black, red, orange, blue, and green solid lines for levels 0, 1, 2, 3, and 4, respectively. $z'$-coordinate is the depth relative to the top of the simulation domain.}
    \label{fig:grid}
\end{figure}

Grid resolution significantly impacts the accuracy of simulations, creating a trade-off between accuracy and computational cost. As discussed in the Introduction, local refinement technique is commonly used to address this challenge. This refinement technique uses increasing grid resolution with decreasing distance from the injection location. In this study, we employ the grid refinement used in Ref.~\cite{wen2023real}. In total, we have five levels of resolutions where level 0 indicates a global level, and four remaining levels (1--4) represent increased resolution near the well (Fig.~\ref{fig:grid}). A higher level indicates a finer resolution and closer distance to the injection location. The details of the discretization are presented in Table S1 of Wen et al.~\cite{wen2023real}. 

As we have five levels of resolutions for predicting pressure buildup and gas saturation, we train one ML model per level (Fig.~\ref{fig:nested}). The output from each level's model is used as an input for the subsequent level's model, in conjunction with reservoir conditions and the injection scheme. Among the five resolution levels, only a model at level 0 does not use predictions from a previous level's, as it is the initial stage of the nested framework. Consequently, this results in a total of five models for predicting pressure buildup. In contrast, there are four models for gas saturation prediction, as no model is trained for gas saturation at level 0.

\subsection{Dataset}\label{sec2.3}

We utilize an open-source dataset provided by Wen et al.~\cite{wen2023real}, obtained from the full physics numerical simulator ECLIPSE (e300)~\cite{eclipse}. The spatial domain of reservoir sampling includes depths ranging from 800 to 4,500 m, with the temperature calculated based on depth and geothermal gradient. For each reservoir, the simulation domain spans a 100-meter depth relative to the top surface, where the upper bound is 800 m. To prevent interactions between different wells, each well is located at least 5,000 m apart, with injection rates ranging from 0.5 to 2.0 MT/y and perforation thicknesses varying from 20 to 100 m. The dataset also considers the variation of the reservoir dip angles from 0 to 2\textdegree.  Stanford Geostatistical Modeling Software (SGeMS)~\cite{Remy2009} was used to produce heterogeneous permeability fields. Table S2 from Wen et al. summarizes the parameters used in generating the dataset, encompassing a wide range of inputs that cover potential scenarios for CO\textsubscript{2} injection into saline formations~\cite{wen2023real}.

The dataset comprises a total of 3,009 reservoir simulations, each featuring between 1 to 4 wells, resulting in a sum of 7,455 local-level simulations across all wells. 80\% of the dataset is used for training, which includes 24 time snapshots, with time intervals exponentially increasing from 1 day to 30 years.

%% file: content/methods.tex
\section{Methods}\label{sec:methods}
While scientific machine learning has seen numerous attempts to apply deep neural operators to two-dimensional (2D) problems, there has been a lack of efforts to model the 3D spatial domain. 
Wen et al.~\cite{wen2023real} initially employed a nested FNO architecture, demonstrating good accuracy for 3D GCS. However, this architecture suffers from high computational costs and discontinuity of the solution in terms of temporal coordinates~\cite{wen2023real, jiang2024fourier}. In response, we propose a nested Fourier-DeepONet, which offers improved performance compared to FNO in terms of prediction accuracy while significantly reducing computational cost.

\subsection{Fourier-DeepONet}\label{sec3.1}
We develop a 3D Fourier-DeepONet based on the 2D Fourier-DeepONet proposed in Refs.~\cite{zhu2023fourier,jiang2024fourier}, which is a hybrid architecture combining a deep operator network (DeepONet)~\cite{Lu2021_DEEPONET} and a Fourier neural operator (FNO)~\cite{Li2020}. DeepONet extends the theorem proposed by Chen and Chen~\cite{TianpingChen1995}, which asserts that nonlinear operator mappings between input and output functions can be learned from data using neural networks. On the other hand, the FNO parameterizes the integral kernel in Fourier space~\cite{Li2020}. FNO computes convolutions in Fourier space rather than physical space to map the input function and the output function discretized on the same equispaced grid.

\begin{figure}[htbp]
    \centering
        \includegraphics[width=0.9\textwidth]{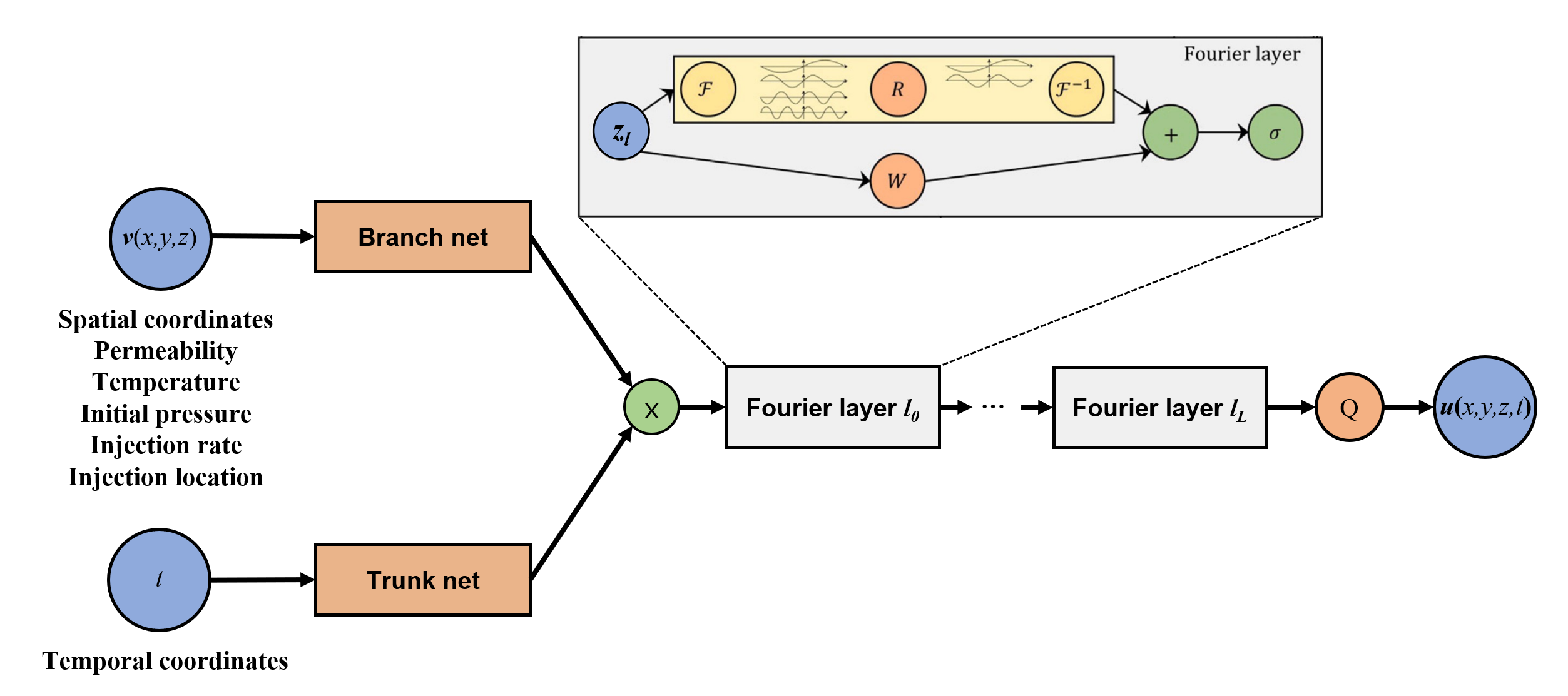}%
    \caption{\textbf{Fourier-DeepONet architecture.} The branch and trunk nets receive two groups of inputs, which are encoded and then decoded by the Fourier layers.} \label{fig:architecture}
\end{figure} 

In Fourier-DeepONet (Fig.~\ref{fig:architecture}), we use DeepONet to encode input through a branch net: permeability map, reservoir temperature, initial pressure, and injection rates. At the same time, the trunk net takes the temporal coordinates $t$ as input.
The outputs of branch net, $\mathbf{b}$, and trunk net, $\mathbf{c}$, are merged through pointwise multiplication: 
\begin{equation*}
    \mathbf{z}_0 = \mathbf{b} \odot \mathbf{c},
\label{eq15}
\end{equation*}
where $\mathbf{z}_0$ represents the output of the branch-trunk merge operation, serving as the input to the first Fourier layer. 

Fourier-DeepONet consists of $L$ Fourier layers which are then applied iteratively starting from the transformed input $\mathbf{z}_0$ on the same mesh as the outputs (pressure buildup, gas saturation). Each Fourier layer performs the following:
\begin{equation*}
    \mathcal{F}^{-1}(\mathcal{R}_l \cdot \mathcal{F}(\mathbf{z}_l)),
\label{eq10}
\end{equation*}
where $\mathcal{F}$ uses 3D Fast Fourier Transform (FFT) and $\mathcal{F}^{-1}$ uses the inverse 3D FFT. And, $\mathbf{z}_l$ denotes the $l$th Fourier layer output. $R_l$ is the weight matrix applied in the Fourier space. Another weight matrix $W_l$ is used for a residual connection to compute the output of the ($j$+1)th Fourier layer output, $\mathbf{z}_{l+1}$, using $\mathbf{z}_l$ as input. Therefore, output of each Fourier layer can be defined as follows:
\begin{equation*}
    \mathbf{z}_{l+1}=\sigma(\mathcal{F}^{-1}(\mathcal{R}_l \cdot \mathcal{F}(\mathbf{z}_l)+W_l \cdot \mathbf{z}_l+b_l),
\label{eq11}
\end{equation*}
where $b_l$ is the bias at $l$th Fourier layer and $\sigma$ is the nonlinear activation function.
Final output of the $L$ Fourier layer is then locally transformed back to the output function dimension parametrized by a fully-connected neural network, $Q$, as
\begin{equation*}
    \textit{\textbf{u}}(\textit{x,y,z}) = Q(\mathbf{z}_L(\textit{x,y,z})).
\label{e12}
\end{equation*}
In this study, 4 Fourier layers are utilized to decode the output. 

Fourier-DeepONet is computationally more efficient than FNO because Fourier-DeepONet uses 3D FFT while FNO uses 4D FFT. This reduction occurs because the output of the trunk net undergoes pointwise multiplication with the branch net output, and thus the final output of Fourier-DeepONet does not include the time dimension. Additionally, Fourier-DeepONet can adopt a batch size for both branch and trunk nets, while FNO is incapable of adjusting the batch size for the time coordinates for training. This further reduces the computational cost compared with using a full time batch size.

\subsection{Nested Fourier-DeepONet}\label{sec3.2}
As discussed in Section \ref{sec2.2}, to predict pressure buildup, we have five levels of grid resolution that result in five different Fourier-DeepONets. For gas saturation prediction, we use four different Fourier-DeepONets for the four resolution levels. We train each Fourier-DeepONet independently at each level for each output and make a nested prediction starting from level 0 up to level 4.

\subsubsection{Networks and resolutions}
Fourier-DeepONet used for each level is denoted as $\mathcal{N}^P_i$, where $i$ in the subscript denotes the level of resolution in Fig.~\ref{fig:nested} for pressure buildup prediction. Similarly, for the prediction of gas saturation, we denote the neural operator as $\mathcal{N}^S_i$ and we only train from level 1 to 4. The spatial domains of Fourier-DeepONets at five different levels are $\Omega_0, \Omega_1, \Omega_2, \Omega_3, \Omega_4$ (Fig.~\ref{fig:grid}). The entire spatial and temporal domain of each Fourier-DeepONet is $D = \Omega_i \times T$, and $T$ denotes the temporal domain up to 30 years. The architecture of each Fourier-DeepONet in nested Fourier-DeepONet, including the output shape of each operation for levels 0 to 4, is shown in Tables~\ref{table:global},~\ref{table:lgr1}, and~\ref{table:lgr2-4} in the Appendix~\ref{appendix:a}.

\subsubsection{Training}
\label{training}
For training $\mathcal{N}^P_0$, we use the reservoir condition and injection scheme as input to the neural operator. For training subsequent levels, we use the true output from the previous level as input in addition to the reservoir condition and injection scheme. Specifically, $\mathcal{N}^P_1$ and $\mathcal{N}^S_1$ use pressure buildup in $\Omega_0$ surrounding each well in the $x$ and $y$ directions with a width and length of 48 km. For the remaining models, $\mathcal{N}^P_i$ and $\mathcal{N}^S_i$ for $i$ = 2--4, the entire ground truth output of the previous level is used for training to predict pressure buildup or gas saturation.

\subsubsection{Inference}
For a reservoir with $n$ wells, the final prediction is combined from $(4 \times n)+1$ network predictions for either pressure buildup or gas saturation prediction. This is because there are four levels of refinement around each injection location, which results in four network predictions per well in addition to the global level prediction. The prediction of pressure buildup at the global level (level 0) around each well will be used as input for the next level (level 1) for pressure buildup and gas saturation per well. Then, each subsequent level $i$, where $i$ = 2--4, will be provided with entire outputs from the previous level $i - 1$. When different levels of resolution are combined to construct the entire spatial domain, the subsequent level replaces the overlapping region.

Although we use ground truth output from the previous level as an input for training purposes, it is not possible to obtain ground truth from the previous level in inference scenarios, which necessitates a sequential prediction as described above. Because the nested architecture is designed in such a way that subsequent levels depend on predictions from the previous level, it is susceptible to accumulating errors. To address this, a fine-tuning step is performed to enhance the generalizability. 

\subsection{Fine-tuning procedure}\label{sec3.3}

We adopt a fine-tuning procedure by adding a random perturbation to the network inputs as introduced in Ref.~\cite{wen2023real}. Specifically for training the subsequent level's model, we perturb the ground truth output from the previous level by adding the randomly sampled model error. For instance, after the training procedure described in Section~\ref{training}, to fine-tune the pressure buildup model at level $i$, the model is further trained using the noised input computed as follows:
\begin{equation*}
    P'_{i-1} = P_{i-1} + \epsilon_{i-1}.
    \label{eq16}
\end{equation*}
$P'_{i-1}$ corresponds to noised ground truth pressure buildup ($P_{i-1}$). To generate the noise, $\epsilon_{i-1}$, we apply the trained model at level $i-1$ to the training dataset and compute the error between the ground truth and the network predictions which results in a set of errors. Then, we randomly sample from the set of errors to obtain the error, $\epsilon_{i-1}$. The same procedure can be applied to fine-tuning gas saturation models. We follow Ref.~\cite{wen2023real} and apply this fine-tuning procedure to $\mathcal{N}^P_1$, $\mathcal{N}^P_4$, $\mathcal{N}^S_1$, and $\mathcal{N}^S_2$. 

\subsection{Details of training and evaluation}\label{sec3.4}

\begin{figure}[htbp]
    \centering
        \includegraphics[width=\textwidth]{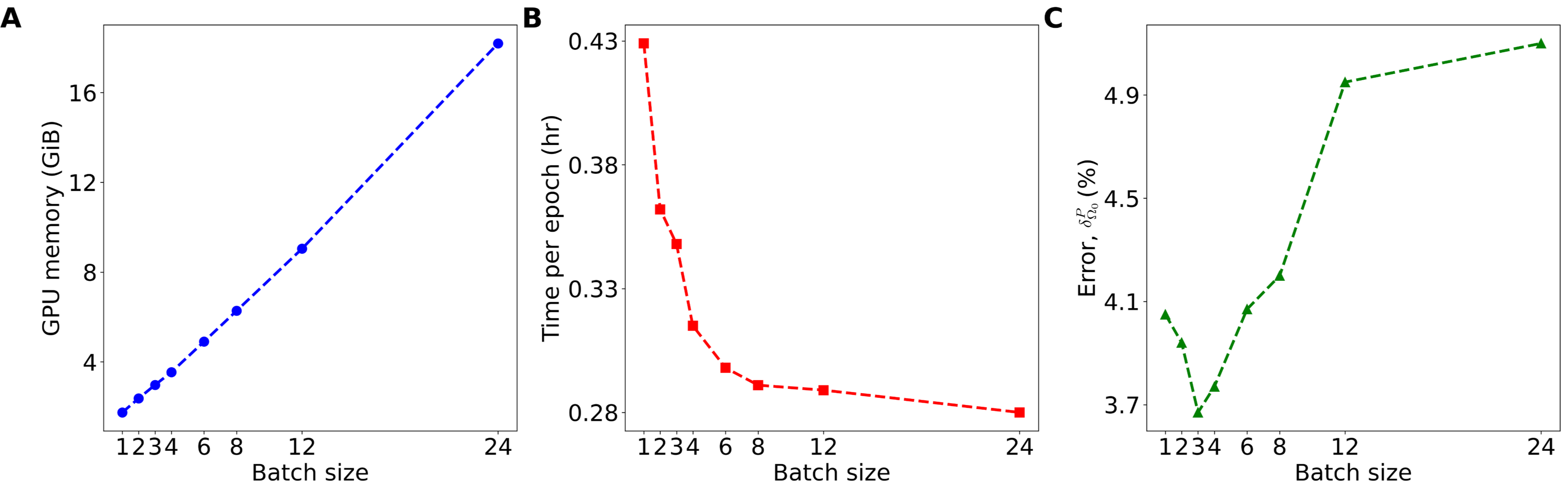}
    \caption{\textbf{Effect of time batch size for Fourier-DeepONet}. (\textbf{A}) GPU memory usage. (\textbf{B}) Training time per epoch. (\textbf{C}) Error, $\delta^{P}_{\Omega_{j}}$, for pressure buildup model at level 0 ($\mathcal{N}^P_0$). Networks are trained for 20 epochs.} \label{fig:batchsize}
\end{figure}

As Fourier-DeepONet separates temporal coordinates through a trunk network, we have an additional option to select their batch size for training. We observe that GPU memory consumption grows linearly (Fig.~\ref{fig:batchsize}A) and training time decreases with respect to the time batch size (Fig.~\ref{fig:batchsize}B) at the global level's pressure buildup model as an example. In this work, we used a batch size of 6 among the 24-time snapshots, as it significantly reduces GPU memory usage without sacrificing model accuracy (Fig.~\ref{fig:batchsize}C) and training time efficiency. 

For the branch net, we use a batch size of 1, identical to that of Wen et al.'s FNO~\cite{wen2023real}. We utilized the $L^2$ relative error for training the nested Fourier-DeepONet. The learning rate is set to 0.001, and the Adam optimizer is employed with a decay rate of 0.9 every two epochs. To evaluate error between the model prediction ($\hat{P}$ and $\hat{S}$) and ground truth ($P$ and $S$), we adopt the same metric as used by Wen et al.~\cite{wen2023real}:
\begin{equation}
    \text{Error of $\mathcal{N}^P_j$  at level }j=0,\dots,4: \quad \delta^{P}_{\Omega_{j}} = \frac{1}{n_{T}n_{\Omega_j}} \sum_{t \in T} \sum_{i \in \Omega_j}\frac{|P_{t,i} - \hat{P}_{t,i}|}{P_{t,max}},
\label{eq:delta_P_j}
\end{equation}
\begin{equation}
    \begin{gathered}
    \text{Error of $\mathcal{N}^S_j$ at level }j=1,\dots,4: \quad
        \delta^{S}_{\Omega_{j}} = \frac{1}{\sum I_{t,i}} \sum_{t \in T} \sum_{i \in \Omega_j}I_{t,i}|S_{t,i} - \hat{S}_{t,i}|.
    \end{gathered}
    \label{eq:delta_S_j}
\end{equation}
where $I_{t,i} = 1$ if $S_{t,i} > 0.01$ or $|\hat{S}_{t,i}| > 0.01$ which is used to prevent overestimation and focus on the error within the plume. Here, $T$ denotes the $n_T$ ( = 24) time snapshots over 30 years, given by $T$ = \{10d, 20d, 30d, 50d, 80d, 110d, 150d, 210d, 280d, 1.0y, 1.3y, 1.7y, 2.2y, 2.8y, 3.6y, 4.6y, 5.9y, 7.5y, 9.4y, 11.9y, 15.0y, 19.0y, 23.9y, 30y\}. ${n_\Omega}_j$ represents the number of cells in the mesh of $\Omega_j$. For pressure buildup error, $\delta^{P}_{\Omega_j}$, the absolute difference between the prediction, $\hat{P}$, and the ground truth, $P$, at each level is normalized by the maximum pressure of the reservoir at each time step, $P_{t,max}$. To evaluate the total error per reservoir, $\delta^{S}_{\Omega}$ or $\delta^{S}_{\Omega}$, we use the following metric:
\begin{equation}
    \text{Total error of }\mathcal{N}^P_0,\mathcal{N}^P_1,\dots, \mathcal{N}^P_4: \quad \delta^{P}_{\Omega} = \frac{1}{n_{T}n_{\Omega}} \sum_{t \in T} \sum_{i \in \Omega}\frac{|P_{t,i} - \hat{P}_{t,i}|}{P_{t,max}},
\label{eq:delta_P}
\end{equation}
\begin{equation}
    \begin{gathered}
        \text{Total error of }\mathcal{N}^S_1,\dots, \mathcal{N}^S_4: \quad \delta^{S}_{\Omega} = \frac{1}{\sum I_{t,i}} \sum_{t \in T} \sum_{i \in \Omega}I_{t,i}|S_{t,i} - \hat{S}_{t,i}|.
    \end{gathered}
    \label{eq:delta_S}
\end{equation}

%% file: content/results.tex
\section{Results}\label{sec:results}

In this section, we highlight the superior computational efficiency of Fourier-DeepONet in terms of both GPU memory consumption and training time. Additionally, we demonstrate the generalization capabilities of the nested Fourier-DeepONet through various experiments, including temporal extrapolation and extrapolation on subsurface conditions and injection schemes. All training experiments were conducted using the NVIDIA A100 80GB GPU.
All codes, implemented using the Python library DeepXDE~\cite{Lu2021deepxde}, and the dataset will be provided in the following GitHub repository: \url{https://github.com/lu-group/nested-fourier-deeponet-gcs-3d}.

\subsection{Computational efficiency}\label{sec4.1}

As discussed in Section~\ref{sec3.1}, the reduced dimension of FFT and flexible batch size for time coordinates result in twice faster training and significant (at least 80\%) reduction in GPU memory consumption (Table~\ref{table:comput_cost}), enabling the training of Fourier-DeepONets even on GPUs with limited memory. Additionally, they lead to a substantial reduction of more than 80\% in the number of trainable parameters, both at global and local resolutions, compared to FNO in Ref.~\cite{wen2023real}.

\begin{table}[htbp]
    \centering
    \caption{\textbf{Computational cost comparison between training nested FNO and nested Fourier-DeepONet}. Fourier-DeepONet uses a time batch size of 6, while FNO is required to use all 24 time batches.}
    \begin{tabular}{lcccc} 
        \toprule
        Model & Level & No. of parameters & GPU memory (GiB) & Training time (hour) \\ \midrule
        \multirow{3}{*}{FNO~\cite{wen2023real}} & Global & 80.3 M & 33.1 & 37.6 \\
        & Local 1 & 150.5 M & 18.5 & 41.7 \\ 
        & Local 2--4 & 150.5 M & 25.4 & 61.3 \\  
        \midrule
        \multirow{3}{*}{Fourier-DeepONet} & Global & 13.1 M & 4.9 & 14.9 \\
        & Local 1& 20.8 M & 3.3 & 20.5 \\
        & Local 2--4 & 20.8 M & 5.0 & 28.3 \\  
        \bottomrule
    \end{tabular}
    \label{table:comput_cost}
\end{table}
\subsection{Pressure buildup prediction}\label{sec4.2}
In addition to the computational efficiency, nested Fourier-DeepONet demonstrates similar or better accuracy for predicting pressure buildup.
We compare the accuracy of Fourier-DeepONets and the state-of-the-art FNOs~\cite{wen2023real} at all levels in Table~\ref{table:accuracy_dp}. As the code for computing the evaluation metrics from Ref.~\cite{wen2023real} was not public, we implemented our own code for Eqs.~\eqref{eq:delta_P_j}--\eqref{eq:delta_S}. For the accuracy of FNO, we include the accuracy from Ref.~\cite{wen2023real} and we also report the accuracy from our own FNOs trained using their public code but evaluated using our evaluation metric code.

Regardless of whether the models are fine-tuned or not, nested Fourier-DeepONet showcases superior performances across all levels compared with nested FNO we trained. Because errors in Table~\ref{table:accuracy_dp} accumulate at each subsequent level, to demonstrate the accuracy of each individual network, we present the test errors using ground truth input in Table~\ref{table:separate} in Appendix~\ref{appendix:b}. For individual networks, Fourier-DeepONet has better accuracy for almost all levels except level 3 for the prediction of pressure buildup. 
 
\begin{table}[htbp]
    \centering
    \caption{\textbf{Accuracy comparison for nested Fourier-DeepONet and nested FNO}. For nested FNOs, we include the errors from Ref.~\cite{wen2023real} and the errors from our own FNOs trained using their public code. Bold font indicates the two smallest errors per row.}
    \begin{tabular}{M{0.13cm}M{1.0cm}M{2.0cm}M{2.0cm}M{2.0cm}M{2.0cm}M{2.0cm}M{2.0cm}} \toprule
        & & \multicolumn{2}{c}{Nested Fourier-DeepONet} & \multicolumn{2}{c}{Nested FNO} & \multicolumn{2}{c}{Nested FNO~\cite{wen2023real}}  \\ \cmidrule(lr){3-4} \cmidrule(lr){5-6} \cmidrule(lr){7-8}
        &  & {Without fine-tuning} & {With fine-tuning} & {Without fine-tuning} & {With fine-tuning} & {Without fine-tuning} & {With fine-tuning} \\ \midrule 
        \multirow{6}{*}{\rotatebox{90}{\small Pressure}} & $\delta^P_\Omega$ & 0.668\% & \textbf{0.565}\% & 0.833\% & 0.625\% & -- & \textbf{0.47}\% \\ \cmidrule(lr){2-8}
        & $\delta^P_{\Omega_0}$ & 0.020\% & \textbf{0.020}\% & 0.025\% & 0.025\% & 0.02\% & \textbf{0.02}\% \\
        & $\delta^P_{\Omega_1}$ & 0.174\% & \textbf{0.173}\% & 0.222\% & 0.233\% & 0.24\% & \textbf{0.16}\% \\ 
        & $\delta^P_{\Omega_2}$ & 0.329\% & \textbf{0.318}\% & 0.400\% & 0.374\% & 0.41\% & \textbf{0.30}\% \\ 
        & $\delta^P_{\Omega_3}$ & 0.557\% & \textbf{0.550}\% & 0.654\% & 0.576\% & 0.68\% & \textbf{0.51}\% \\ 
        & $\delta^P_{\Omega_4}$ & 1.433\% & \textbf{1.110}\% & 1.801\% & 1.205\% & 1.88\% & \textbf{0.82}\% \\ \midrule
        \multirow{5}{*}{\rotatebox{90}{\small Saturation}} & $\delta^S_\Omega$ & \textbf{1.518}\% & \textbf{1.461}\% & 1.663\% & 1.622\% & -- & 1.79\% \\ \cmidrule(lr){2-8}
        & $\delta^S_{\Omega_1}$ & \textbf{1.331}\% & \textbf{1.290}\% & 1.650\% & 1.419\% & 1.55\% & 1.39\% \\ 
        & $\delta^S_{\Omega_2}$ & \textbf{1.675}\% & \textbf{1.569}\% & 1.936\% & 1.802\% & 1.93\% & 1.91\% \\ 
        & $\delta^S_{\Omega_3}$ & \textbf{1.649}\% & \textbf{1.583}\% & 1.888\% & 1.737\% & 2.00\% & 1.77\% \\ 
        & $\delta^S_{\Omega_4}$ & \textbf{1.439}\% & \textbf{1.380}\% & 1.563\% & 1.545\% & 1.95\% & 1.82\% \\ \bottomrule
    \end{tabular}
    \label{table:accuracy_dp}
\end{table}

\begin{figure}[htbp]
    \centering
        \includegraphics[width=\textwidth]{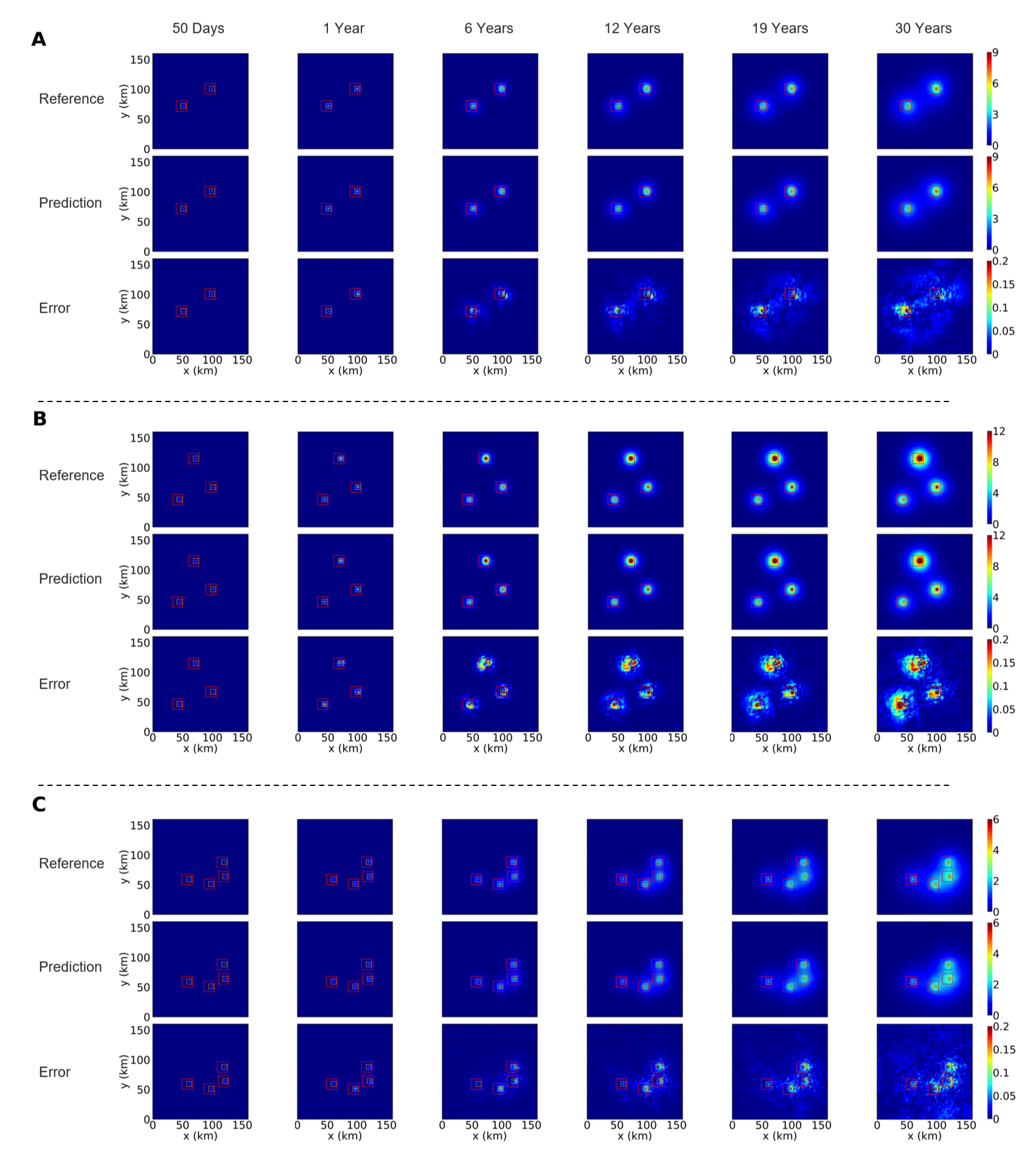}
    \caption{\textbf{Examples of nested Fourier-DeepONet pressure buildup prediction for three different reservoirs on the top surface of the reservoir.} (\textbf{A}) 2 injection wells. (\textbf{B}) 3 injection wells. (\textbf{C}) 4 injection wells. } \label{fig:dp_example1}
\end{figure} 

\begin{figure}[htbp]
    \centering
        \includegraphics[width=\textwidth]{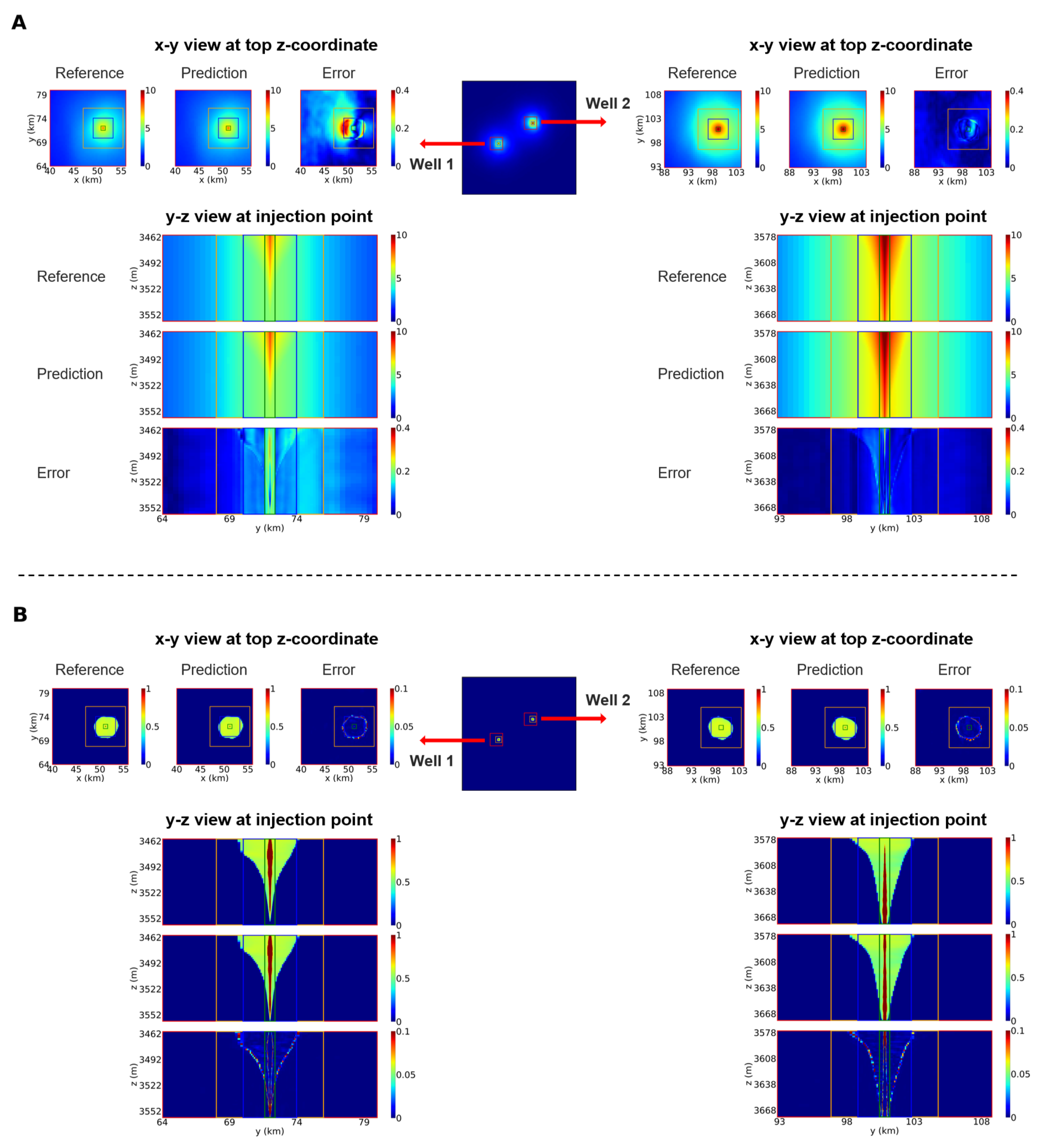}
    \caption{\textbf{Nested Fourier-DeepONet prediction at local refinements corresponding to the first example of Fig.~\ref{fig:dp_example1}A at 30 years.} (\textbf{A}) Pressure buildup. (\textbf{B}) Gas saturation.}
    \label{fig:lgr_all}
\end{figure}

We show pressure buildup predictions and errors ($xy$-view at the top layer) of three randomly selected reservoirs with 2--4 wells at 6 different time snapshots, $T$ = \{50d, 1.0y, 7.5y, 19.0y, 23.9y, and 30y\} in Fig.~\ref{fig:dp_example1}. At early times where there is no significant pressure gradient, the error is the lowest. The error increases with time as the injection of CO$_2$ results in the buildup of pressure. For the pressure buildup of the first example (Fig.~\ref{fig:dp_example1}A), both cross-sectional $xy$-view at the top layer and $xz$-view around the injection point (level 1--4) are shown in Fig.~\ref{fig:lgr_all}A. By employing locally refined models, the errors are significantly reduced through predictions at refined resolutions.

\subsection{Gas saturation prediction}\label{sec4.3}
For gas saturation network models, the accuracy of nested Fourier-DeepONet and nested FNO is also shown in Table~\ref{table:accuracy_dp}. As mentioned in Section~\ref{sec3.4}, we do not train the gas saturation model at the global level. As we were able to achieve better accuracy at the global level ($\delta^P_{\Omega_0}$) and level 1 ($\delta^S_{\Omega_1}$) for individual prediction (Table~\ref{table:separate}), the cumulative error arising from earlier levels resulted in lower accuracy with Fourier-DeepONets for both fine-tuned models and those without tuning. This highlights the importance of achieving robust accuracy at earlier levels when using a nested framework. 

As shown in Fig.~\ref{fig:lgr_all}B, the CO$_2$ plume footprint is contained within the domain of level 1 even at 30 years of CO$_2$ injection. The error tends to be the largest at the boundary of the footprint in addition to the injection location. 

\subsection{Generalization and extrapolation capability of nested Fourier-DeepONet}\label{sec4.4}

To further investigate the generalizability of nested Fourier-DeepONet on unseen datasets, we conducted experiments to assess its performance under different extrapolation scenarios in terms of reservoir conditions (Sections \ref{sec4.4.1} and \ref{sec4.4.2}), injection schemes (Section \ref{sec4.4.3}), and temporal coordinates (Section \ref{sec4.4.4}). For temporal extrapolation, we train models based on the first 21 time snapshots out of 24 while using all training samples.


\begin{figure}[htbp]
    \centering
    \includegraphics[width=\textwidth]{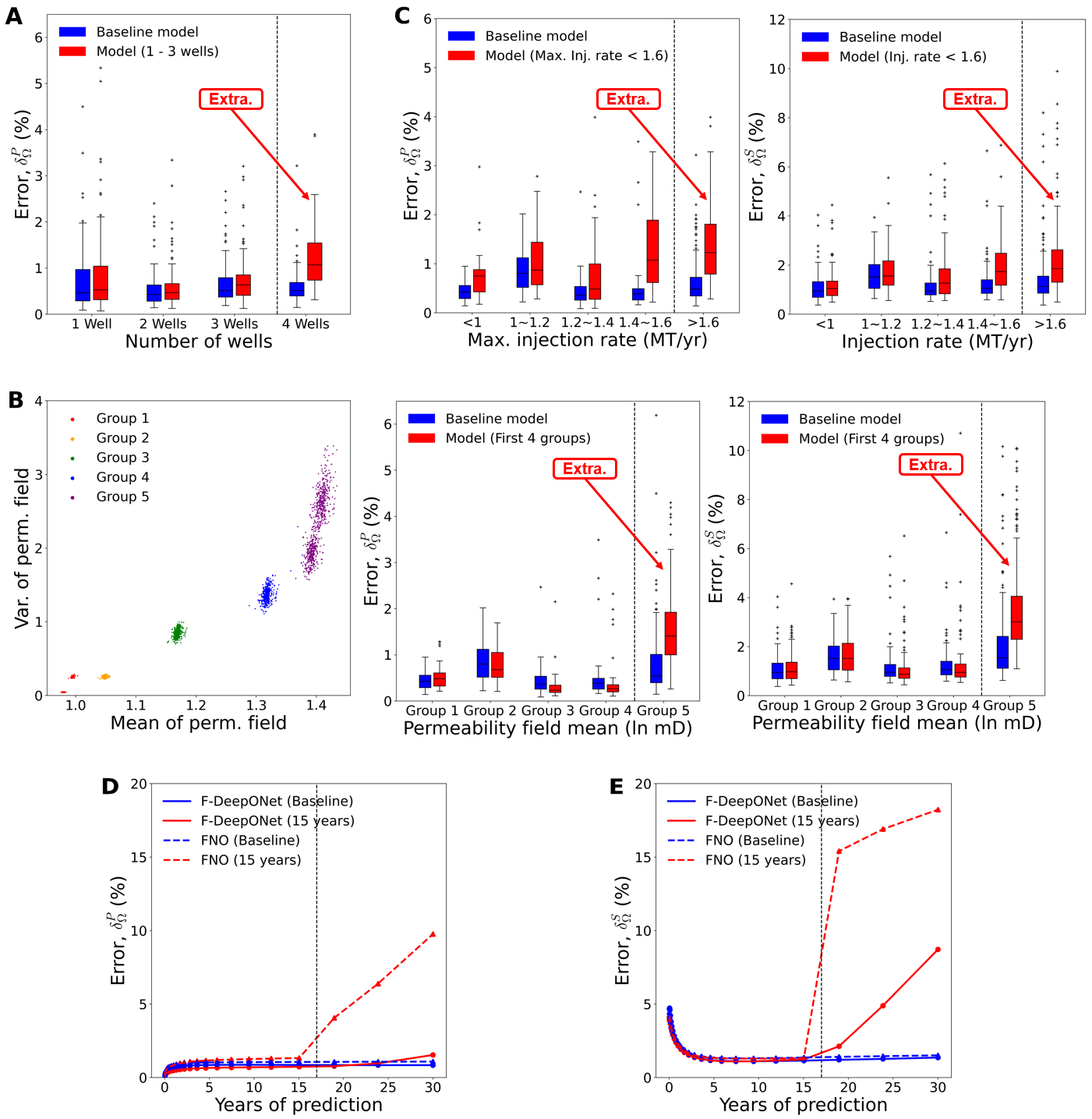}
    \caption{\textbf{Interpolation and extrapolation errors for different reservoir conditions, injection schemes, and temporal coordinates.} (\textbf{A}, \textbf{B}, and \textbf{C}) Interpolation and extrapolation errors of pressure buildup for different conditions. (A) Number of wells. (B) Permeability field magnitude. (C) Maximum injection rate. (\textbf{D} and \textbf{E}) Interpolation and extrapolation errors of gas saturation for different conditions. (D) Permeability field magnitude. (E) Injection rate. (\textbf{F} and \textbf{G}) Comparison between nested Fourier-DeepONet and nested FNO for temporal extrapolation. (F) Errors of pressure buildup. and (G) Errors of gas saturation.} \label{fig:extra_all}
\end{figure} 

\subsubsection{Number of wells}\label{sec4.4.1}
Our dataset contains reservoirs with 1--4 injection wells. Here, we investigate the extrapolation capabilities of Fourier-DeepONet on the number of wells. Specifically, reservoirs containing 1--3 wells (1806 samples) were used to train Fourier-DeepONet at level 0, $\mathcal{N}^P_0$. Then, we evaluate their prediction accuracy on reservoirs containing 4 wells for pressure buildup prediction, which we refer to as extrapolation. We did not separately train local level (1--4) models for this scenario because the local level models are not subject to the number of wells.

We observe comparable errors for predictions on reservoirs with 1--3 wells (i.e. interpolation) between the baseline model and the model trained with 1--3 wells (Fig.~\ref{fig:extra_all}A). The baseline model refers to the nested Fourier-DeepONet without fine-tuning in Table~\ref{table:accuracy_dp}, which has been trained on the entire dataset. For reservoirs with 4 wells, the third quartile of error in predicting pressure buildup using the extrapolation model is below 2\%.  This result suggests that the nested Fourier-DeepONet demonstrates strong generalization capabilities across different well configurations.

\subsubsection{Permeability field}\label{sec4.4.2}
We assess the extrapolation capability of Fourier-DeepONet on the permeability field by dividing the dataset into five clusters based on the average value of the 3D permeability field in the reservoirs. In the dataset, the permeability fields are well clustered into five groups when projected into a 2D space of mean and variance values (Fig.~\ref{fig:extra_all}B). The cutoff values separating the clusters, from Group 1 to Group 5, are 1.02, 1.1, 1.25, and 1.377 ln mD, as shown in different colors in Fig.~\ref{fig:extra_all}B. For this experiment, we select samples from the first four groups for training, which represent 55\% (1637 samples) of the original training dataset $\mathcal{N}^P_0$.

For both pressure buildup and gas saturation prediction, the errors between the baseline model and the extrapolation model are comparable in the interpolation regime (Group 1 to Group 4). Despite the observed error increase in the extrapolation regime (Group 5), the median error for pressure buildup prediction remains below 2\%, and the error for gas saturation prediction stays below 5\% (Fig.~\ref{fig:extra_all}B). Because samples with high permeability field mean have larger variances, and having a highly permeable field leads to easier migration of CO$_2$ through the media, it can become more difficult for the model to make accurate predictions under such subsurface conditions. This implies that it is more efficient to allocate computing resources to obtaining datasets of permeability fields of a high magnitude. This analysis offers further insight into the data generation process, indicating that training the models with a broader range of permeability fields, particularly at higher permeabilities, would enhance overall accuracy with minimal tradeoffs regarding lower prediction errors at a lower permeability regime.

\subsubsection{Injection rate}\label{sec4.4.3}
Next, we conduct an extrapolation study based on injection rates. We differentiate the samples by their maximum injection rate for pressure buildup models, but we look at each injection rate to differentiate the samples for gas saturation models. Samples with multiple wells inherently have a higher probability of having a larger maximum injection rate, so we use less number of samples for training pressure buildup models (1155 for $\mathcal{N}^P_{0}$ and 2425 for $\mathcal{N}^P_{1-4}$) than gas saturation models (4325 for $\mathcal{N}^s_{1-4}$). The pressure buildup extrapolation models are trained with samples with maximum injection rates up to 1.6 MT/year at the global level and the corresponding samples are selected for training at local levels. On the other hand, we only consider the injection rate for each well for training gas saturation models. 

As shown in Fig.~\ref{fig:extra_all}C, we observe a moderate increase in error when predicting pressure buildup for maximum injection rates above 1.6 MT/yr per reservoir. This increase is not only due to extrapolation errors but also the reduced number of training samples ($<$ 40\% of the total samples) available at injection rates below 1.6 MT/yr. Notably, the error increase observed when using a model trained on injection rates up to 1.6 MT/yr for predicting samples in the 1.4--1.6 MT/yr range (interpolation regime) is comparable to that seen in the extrapolation regime relative to a baseline model. This suggests that the contribution of extrapolation errors may be minimal, as a similar error increase is observed within the interpolation regime. Similarly, the error in gas saturation predictions increases to a similar extent for samples with injection rates between 1.4 and 1.6 MT/yr and those above 1.6 MT/yr. This trend indicates that the extrapolation error in the gas saturation model trained on injection rates below 1.6 MT/yr is also likely to be small.

\subsubsection{Extrapolation in time}\label{sec4.4.4}
The final experiment involves a temporal extrapolation. Instead of training a model with a reduced number of reservoir samples as in Sections~\ref{sec4.4.1} to~\ref{sec4.4.3}, we train models with a limited number of time snapshots (21 out of 24) up to 15 years to assess their generalization capabilities up to 30 years. We train both a nested Fourier-DeepONet and a nested FNO to demonstrate the superior performance of a nested Fourier-DeepONet in temporal extrapolation. 

Nested Fourier-DeepONet, trained up to 15 years, exhibits a similar error in the interpolation regime as the baseline model, but only slightly higher errors in the extrapolation regime (Fig.~\ref{fig:extra_all}D). The rise in error using a nested Fourier-DeepONet is relatively moderate, especially compared to FNO. This is because the use of a trunk net from Fourier-DeepONet allows the solution to be continuous over time, rather than using a separate channel for each time step like FNO. For example, nested Fourier-DeepONet that has been trained up to 15 years extrapolated on 30 years with less than 1\% error increase for pressure buildup prediction compared with a baseline model, which is a significant improvement over FNO. Although we observe a moderate error increase in gas saturation prediction (Fig.~\ref{fig:extra_all}E), Fourier-DeepONet still exhibits roughly a 10\% decrease in error compared to FNO. This experiment enables us to assess the model's performance in the temporal extrapolation regime.

%% file: content/conclusions.tex
\section{Conclusions}\label{sec:conclusion}

In this work, we have developed a nested Fourier-DeepONet framework for simulating a 3D GCS over 30 year period. We also have demonstrated its superior performance over nested FNO not only in accuracy but also in computational efficiency. Fourier-DeepONet is faster in training compared to FNO, thanks to the reduced number of trainable parameters and lower consumption of GPU memory achieved by selecting a smaller time batch size. 

Previous 2D GCS experiments by Jiang et al.~\cite{jiang2024fourier} in temporal interpolation have shown better performance compared to FNO, as the trunk net guarantees continuous prediction with respect to time. Building upon this, our extrapolation experiments have further revealed that the nested Fourier-DeepONet exhibits good extrapolation capability across various scenarios (i.e., number of wells, permeability field, injection rate, time coordinates). This extrapolation analysis provides valuable insight into the maximization of model performance from a limited dataset. 


Currently, there are still some limitations to the application of deep neural operators to real-world problems. First, since the current method relies solely on data, the size of the dataset significantly influences the predictive capability of the neural operators. Although deep neural operators demonstrate good generalizability, subsurface conditions can vary widely between sites. To address this challenge, one common approach is to incorporate physics-informed loss functions during neural network training. Another limitation of the current nested framework is the need to train a model for each level, which can be time-consuming. Future advancements could involve developing architectures capable of handling non-equispaced meshes with varying grid points and resolutions. For example, a deep neural operator should be able to train and predict in reservoirs with varying well configurations, involving different numbers of cells.

%% file: content/appendix.tex
\section{Fourier-DeepONet architecture} 
\label{appendix:a}

The network architectures of Fourier-DeepONets for all levels mentioned in Section~\ref{sec3.2} are shown in Tables~\ref{table:global},~\ref{table:lgr1}, and~\ref{table:lgr2-4}.

\begin{table}[htbp]
    \centering
    \caption{\textbf{Global Fourier-DeepONet architecture.}}
    \begin{tabular}{cccc} \toprule
         &  &  {Operation}& {Output shape}\\ \midrule
         Branch net &  &  Padding(8), Linear & (C, 116, 116, 21, 32)\\
         Trunk net &  &  Linear & (T, 32) \\ 
         Merge operation &  &  Point-wise multiplication & (C $\times$ T, 116, 116, 21, 32) \\ \midrule
         \multirow{8}{5em}{Merge net} & Fourier 1 & Fourier3d/Conv1d/Add/GELU & (C $\times$ T, 116, 116, 21, 32)\\ 
         & Fourier 2 & Fourier3d/Conv1d/Add/GELU & (C $\times$ T, 116, 116, 21, 32)\\   
         & Fourier 3 & Fourier3d/Conv1d/Add/GELU & (C $\times$ T, 116, 116, 21, 32)\\  
         & Fourier 4 & Fourier3d/Conv1d/Add/GELU & (C $\times$ T, 116, 116, 21, 32)\\  
         & De-padding & De-padding(8) & (C $\times$ T, 100, 100, 5, 32)\\   
         & Projection 1 & Linear & (C $\times$ T, 100, 100, 5, 128)\\
         & Projection 2 & Linear & (C $\times$ T, 100, 100, 5, 1)\\   
         & Reshape & - & (C, T, 100, 100, 5) \\ \bottomrule
    \end{tabular}
    \label{table:global}
\end{table}

\begin{table}[htbp]
    \centering
    \caption{\textbf{LGR1 Fourier-DeepONet architecture.}}
    \begin{tabular}{cccc} \toprule
         &  &  {Operation}& {Output shape}\\ \midrule
         Branch net &  &  Padding(8), Linear & (C, 56, 56, 41, 36)\\
         Trunk net &  &  Linear & (T, 36) \\ 
         Merge operation &  &  Point-wise multiplication & (C $\times$ T, 56, 56, 41, 36) \\ \midrule
         \multirow{8}{5em}{Merge net} & Fourier 1 & Fourier3d/Conv1d/Add/GELU & (C $\times$ T, 56, 56, 41, 36)\\ 
         & Fourier 2 & Fourier3d/Conv1d/Add/GELU & (C $\times$ T, 56, 56, 41, 36)\\   
         & Fourier 3 & Fourier3d/Conv1d/Add/GELU & (C $\times$ T, 56, 56, 41, 36)\\  
         & Fourier 4 & Fourier3d/Conv1d/Add/GELU & (C $\times$ T, 56, 56, 41, 36)\\  
         & De-padding & De-padding(8) & (C $\times$ T, 40, 40, 25, 36)\\   
         & Projection 1 & Linear & (C $\times$ T, 40, 40, 25, 144)\\   
         & Projection 2 & Linear & (C $\times$ T, 40, 40, 25, 1)\\   
         & Reshape & - & (C, T, 40, 40, 25) \\ \bottomrule
    \end{tabular}
    \label{table:lgr1}
\end{table}

\begin{table}[htbp]
    \centering
    \caption{\textbf{LGR2--4 Fourier-DeepONet architecture.}}
    \begin{tabular}{cccc} \toprule
         &  &  {Operation}& {Output shape}\\ \hline
         Branch net &  &  Padding(8), Linear & (C, 56, 56, 66, 36)\\
         Trunk net &  &  Linear & (T, 28) \\ 
         Merge operation &  &  Point-wise multiplication & (C $\times$ T, 56, 56, 66, 36) \\ \hline
         \multirow{8}{5em}{Merge net} & Fourier 1 & Fourier3d/Conv1d/Add/GELU & (C $\times$ T, 56, 56, 66, 36)\\ 
         & Fourier 2 & Fourier3d/Conv1d/Add/GELU & (C $\times$ T, 56, 56, 66, 36)\\   
         & Fourier 3 & Fourier3d/Conv1d/Add/GELU & (C $\times$ T, 56, 56, 66, 36)\\  
         & Fourier 4 & Fourier3d/Conv1d/Add/GELU & (C $\times$ T, 56, 56, 66, 36)\\  
         & De-padding & De-padding(8) & (C $\times$ T, 40, 40, 50, 36)\\   
         & Projection 1 & Linear & (C $\times$ T, 40, 40, 50, 144)\\   
         & Projection 2 & Linear & (C $\times$ T, 40, 40, 50, 1)\\   
         & Reshape & - & (C, T, 40, 40, 50) \\ \bottomrule
    \end{tabular}
    \label{table:lgr2-4}
\end{table}

\newpage

\section{Prediction accuracy for individual networks}
\label{appendix:b}

Table~\ref{table:separate} shows the test accuracy of the nested Fourier-DeepONet and nested FNO models, where ground truth outputs from the previous level are used for $\mathcal{N}^P_i$ and $\mathcal{N}^S_i$, with $i$ = 1--4. Although this approach is not feasible in practice, it allows us to assess how well the model was trained at each individual level. Table~\ref{table:accuracy_dp} corresponds to the ``sequential'' prediction accuracy, while Table~\ref{table:separate} represents the ``separate'' prediction accuracy, as presented in Table 7 of Ref.~\cite{wen2023real}.
\begin{table}[htbp]
    \centering
    \caption{\textbf{Accuracy comparison for nested FNO and nested Fourier-DeepONet}. The errors are computed using the test dataset both in the entire domain and each subdomain. Separate refers to the model prediction using the ground truth from the previous level's output as the input.}
    \begin{tabular}{M{0.03cm}M{1.2cm}M{3.5cm}M{3.5cm}M{3.5cm}} \toprule
        & {Domain} & {Nested Fourier-DeepONet} & {Nested FNO} & {Nested FNO~\cite{wen2023real}} \\ \midrule
        \multirow{6}{*}{\rotatebox{90}{\small Pressure}} & $\delta^P_\Omega$ & \textbf{0.069}\% & 0.080\% & - \\ \cmidrule(lr){2-5}
        & $\delta^P_{\Omega_0}$ & \textbf{0.020}\%  & 0.025\% & 0.02\% \\
        & $\delta^P_{\Omega_1}$ & \textbf{0.039}\%  & 0.061\% & 0.10\% \\ 
        & $\delta^P_{\Omega_2}$ & \textbf{0.044}\% & 0.046\% & 0.16\% \\ 
        & $\delta^P_{\Omega_3}$ & 0.059\%  & \textbf{0.058}\% & 0.14\% \\ 
        & $\delta^P_{\Omega_4}$ & \textbf{0.128}\%  & 0.148\% & 0.45\% \\ \midrule
        \multirow{5}{*}{\rotatebox{90}{\small Saturation}} & $\delta^S_\Omega$ & 0.486\% & \textbf{0.461}\% & -\\ \cmidrule(lr){2-5}
        & $\delta^S_{\Omega_1}$ & \textbf{1.073}\% & 1.271\% & 1.27\% \\ 
        & $\delta^S_{\Omega_2}$ & 0.733\% & \textbf{0.667}\% & 1.00\% \\ 
        & $\delta^S_{\Omega_3}$ & 0.562\% & \textbf{0.480}\% & 0.61\% \\ 
        & $\delta^S_{\Omega_4}$ & 0.478\% & \textbf{0.455}\% & 0.74\% \\ \bottomrule
    \end{tabular}
    \label{table:separate}
\end{table}